%% file: scis_paper.tex
\documentclass{SCIS2024}

\usepackage{caption}
\usepackage{multirow}
\usepackage{array}
\usepackage{threeparttable}
\usepackage{rotating}
\usepackage{cleveref}
\usepackage{colortbl}
\usepackage{subfig}
\usepackage{subfloat}
\usepackage{xcolor}
\definecolor{purple}{RGB}{157, 125, 209}
\definecolor{blue}{RGB}{122,174,221}
\definecolor{red}{RGB}{243,166,114}
\definecolor{green}{RGB}{112, 173, 71}
\definecolor{yellow}{RGB}{197,90,17}
\definecolor{grey}{rgb}{0.9,0.9,0.9}
\crefname{section}{§}{§§}
\Crefname{section}{§}{§§}

\begin{document}
    \ArticleType{RESEARCH PAPER}
    \Year{2024}
    \Month{}
    \Vol{}
    \No{}
    \DOI{}
    \ArtNo{}
    \ReceiveDate{}
    \ReviseDate{}
    \AcceptDate{}
    \OnlineDate{}

\title{ChemDFM-X: Towards Large Multimodal Model for Chemistry}{ChemDFM-X: Towards Large Multimodal Model for Chemistry}


\author[1]{Zihan Zhao}{}
\author[2]{Bo Chen}{}
\author[1,2]{Jingpiao Li}{}
\author[1,2]{Lu Chen}{chenlusz@sjtu.edu.cn}
\author[2]{Liyang Wen}{}
\author[1,2]{\\Pengyu Wang}{}
\author[1]{Zichen Zhu}{}
\author[1]{Danyang Zhang}{}
\author[2]{Ziping Wan}{}
\author[1]{\\Yansi Li}{}
\author[2]{Zhongyang Dai}{}
\author[2]{Xin Chen}{mail.xinchen@gmail.com}
\author[1,2]{Kai Yu}{kai.yu@sjtu.edu.cn}

\AuthorMark{Zihan Zhao}

\AuthorCitation{Zihan Zhao, Jingpiao Li, Bo Chen, et al}

\contributions{Zihan Zhao and Bo Chen have the same contribution to this work.}

    \address[1]{X-LANCE Lab, Department of Computer Science and Engineering\\ MoE Key Lab of Artificial Intelligence, SJTU AI Institute\\ Shanghai Jiao Tong University, Shanghai {\rm 200240}, China}
\address[2]{Suzhou Laboratory, Suzhou {\rm 215123}, China}

    \abstract{
    Rapid developments of AI tools are expected to offer unprecedented assistance to the research of natural science including chemistry. However, neither existing unimodal task-specific specialist models nor emerging general large multimodal models (LMM) can cover the wide range of chemical data modality and task categories. To address the real demands of chemists, a cross-modal Chemical General Intelligence (CGI) system, which serves as a truly practical and useful research assistant utilizing the great potential of LMMs, is in great need. In this work, we introduce the first \textbf{Cross}-modal \textbf{D}ialogue \textbf{F}oundation \textbf{M}odel for \textbf{Chem}istry (\textbf{ChemDFM-X}). Diverse multimodal data are generated from an initial modality by approximate calculations and task-specific model predictions. This strategy creates sufficient chemical training corpora, while significantly reducing excessive expense, resulting in an instruction-tuning dataset containing 7.6M data. After instruction finetuning, ChemDFM-X is evaluated on extensive experiments of different chemical tasks with various data modalities. The results demonstrate the capacity of ChemDFM-X for multimodal and inter-modal knowledge comprehension. ChemDFM-X marks a significant milestone toward aligning all modalities in chemistry, a step closer to CGI.
    }


    \maketitle


\input{1.introduction}
\input{2.related_work}
\input{3.chemdfm-x}
\input{4.evaluation}

\input{5.conclusion}

\bibliographystyle{unsrt}
\bibliography{custom}

\end{document}

%% file: 1.introduction.tex
\section{Introduction}
Chemistry, as a naturally multimodal subject of science, plays a crucial role in various vital fields such as pharmaceutical research and material manufacturing. Therefore, research on AI for chemistry has garnered increasing attention. Despite the rapid development, most of the chemical AI models today mainly focus on single tasks with unimodal input~\cite{chemberta,chemformer,shi2023relm,chen2023towards}. However, chemical data covers a wide range of modalities spanning from text description and molecular structure to image and spectrum, and chemical tasks take various forms ranging from property prediction to retrosynthesis. Although these unimodal specialist models can achieve state-of-the-art~(SOTA) performances in their individual tasks, they inherently can not handle tasks even slightly different from their own, or their respective tasks when there is a slight alteration in the input modality. Therefore, the practical utility and assistance of these models in research and manufacturing are limited.

Nowadays, large language models~(LLMs) and large multimodal models~(LMMs)~\cite{mplugdocowl, internlmxcomposer,liu2024improved,gpt-4v,qwenvl,gemini} have achieved impressive performance in a number of challenging fields such as natural image inference~\cite{gemini,liu2024improved,gpt-4v}, document analysis~\cite{mplugdocowl, internlmxcomposer}, and medical image reasoning~\cite{tu2023generalist,yang2024advancing}. Therefore, LMMs show great potential for building a cross-modal Chemical General Intelligence~(CGI) system~\cite{zhao2024chemdfm}. However, most of the previous LMMs only focus on one non-text modality. Given the diversity of chemical modalities and the frequent co-occurrence of different modalities in practice, a single LMM that can handle multiple modalities is needed for CGI to truly meet the requirements of chemists. 

\begin{figure}[t]
        \centering
        \includegraphics[width=\textwidth,trim=20 50 20 50,clip]{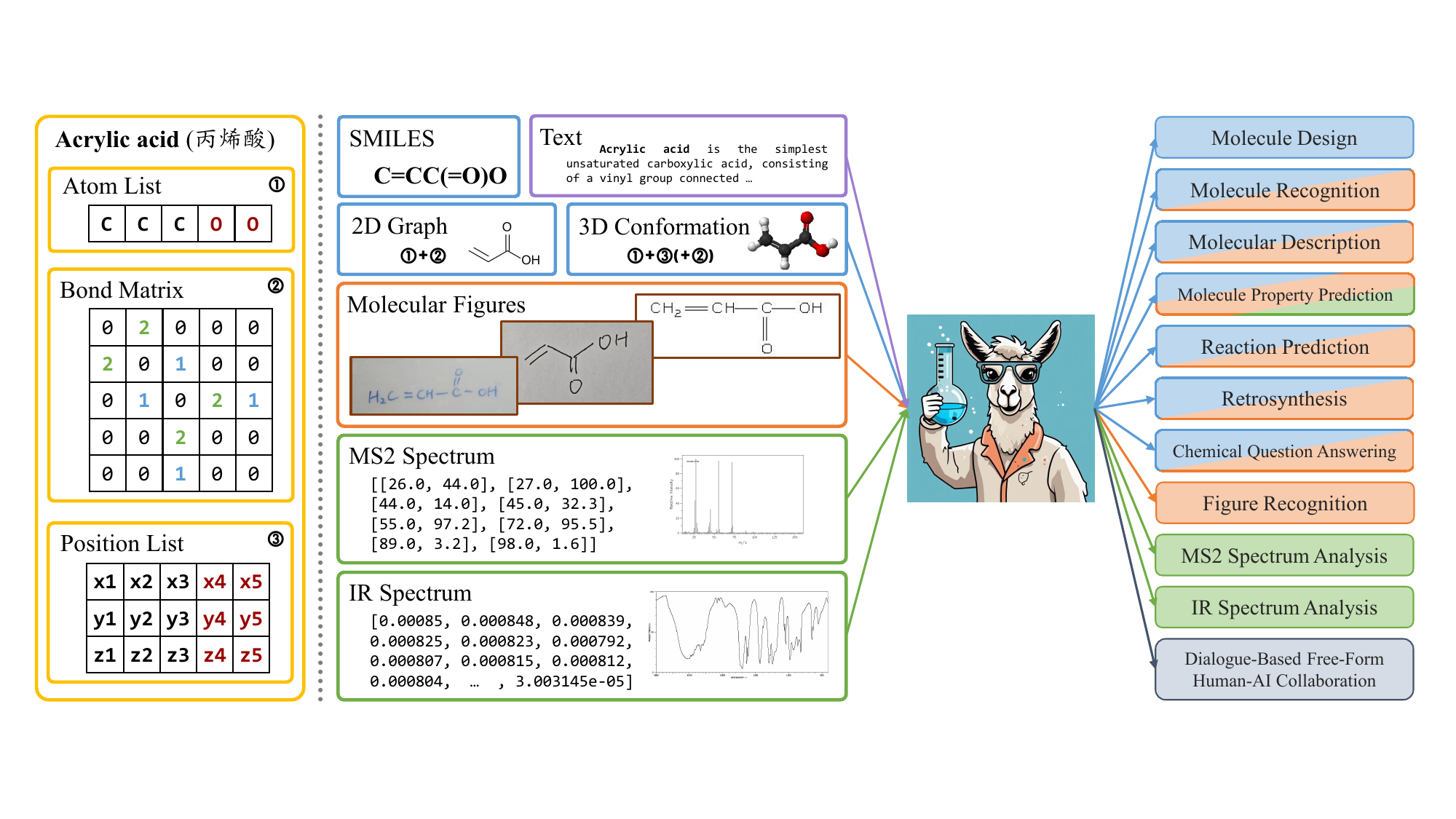}
        \caption{The overview of ChemDFM-X. The different modalities involved in chemical tasks are distinguished by colors. Structural modalities are marked with \textcolor{blue}{\bf blue}, figures are marked with \textcolor{red}{\textbf{orange}}, and spectra are marked with \textcolor{green}{\textbf{green}}. The text modality is marked with \textcolor{purple}{\textbf{purple}} in the input part and omitted in the output part. The dialogue-based free-form human-AI collaboration may involve any feasible modalities and is marked with \textcolor{gray}{\textbf{gray}}. For a detailed introduction to these modalities, please refer to Section~\ref{sec:overview}.}
        \label{figure:overview}
    \end{figure}

In this work, we detail our progress toward such a cross-modal chemical LLM and propose \textbf{ChemDFM-X}, a Cross-modal Dialogue Foundation Model for Chemistry that can comprehend and interpret data of various chemical modalities and fulfill many downstream tasks with the same set of model weights. As shown in Figure~\ref{figure:overview}, ChemDFM-X takes advantage of the pre-trained parameters of the ChemDFM~\cite{zhao2024chemdfm} model and is continuously trained on various multi-modality data.
Specifically, besides text and SMILES\footnote{short for Simplified Molecular-Input Line-Entry System, a linear representation of chemical molecule structures} modalities already learned by ChemDFM, we choose five typical modalities that are both representative and meaningful in the field of chemistry.

One of the key challenges to achieving this goal is the absence of sufficient modality-aligned data. To address this problem, we propose to supplement data in other modalities by converting from SMILES. Noting that large-scale data are difficult to acquire, especially for the characterization modalities such as tandem mass spectra~(MS2) and infrared spectra~(IR), owing to the excessive expenditure of both experiments and quantum chemical calculation, we leverage simplified approximate calculation and numerical model prediction to obtain sub-optimal yet proximal results. In this way, we finally generate a multi-modal instruction-tuning dataset containing 7.6M cross-modality data from 1.3M seed SMILES.

Benefiting from the instruction-tuning dataset, ChemDFM-X possesses the capabilities to comprehend and infer over various modalities including molecular graphs, conformations, images, and spectra.
To demonstrate the prowess of ChemDFM-X, we conduct extensive experiments regarding the newly added modalities. 
The results demonstrate the strong capacity of ChemDFM-X for comprehending multi-modality input and exploiting inter-modality knowledge. Compared to conventional specialist models and LMMs which only enable one or none of the chemical modalities, ChemDFM-X manages to handle most common modalities well and agilely leverages the knowledge of chemical materials and reactions learned from all the modalities to solve various practical chemical tasks with superior performances.
To the best of our knowledge, ChemDFM-X is the first demonstration of a cross-modality chemical general intelligence system that can interpret chemical data of multiple different modalities with the same sets of parameters while handling a wide variety of tasks.

%% file: 2.related_work.tex
\section{Related Work}

AI for Chemistry has long been an active research area and has recently garnered increasing attention~\cite{molformer, chemberta, honda2019smiles,graphcl, xia2023molebert, MGSSL, ying2021transformers,Denoising, 3dEMGP, zhou2023unimol, chen2023towards,shi2023relm,sagawa2023reactiont5, fang2023mol}. Due to the foundational role of molecules in the chemical world, the AI for Chemistry research primarily revolves around molecules~\cite{molformer, chemberta, honda2019smiles,graphcl, xia2023molebert, MGSSL, ying2021transformers,Denoising, 3dEMGP, zhou2023unimol} and reactions~\cite{chen2023towards,shi2023relm,sagawa2023reactiont5} which are composed of molecules. To better embed molecules into AI models, researchers mainly exploit three different molecular representations: 1) SMILES notations~\cite{molformer,chemberta,honda2019smiles}, 2) 2D molecular graph~\cite{graphcl,xia2023molebert,MGSSL,ying2021transformers}, and 3) 3D molecular conformations~\cite{Denoising, 3dEMGP, zhou2023unimol}. In recent years, many works have incorporated different molecular representations together for better performance. 
DMP~\cite{zhu2021dualview} encodes molecular SMILES and molecular graphs separately using a transformer and graph neural network. Through masked atom modeling and contrastive learning in both modalities, DMP pretrains the model to support inputs from both SMILES and molecular graphs. SGGRL~\cite{Wang2024MultiModalRL} uses sequence-based, graph-based, and geometry-based encoders to obtain molecular representations containing diverse modal information. Then a readout layer captures crucial information from various modalities to acquire molecule-level representations, and a fusion layer integrates different modal representations using attention mechanisms to derive the ultimate molecular representation.
Although the above task-specific specialist models are able to achieve advanced performance on their corresponding tasks, they suffer from poor task generalization capabilities, and can not be applied to other tasks. Therefore, their practical utility and value is limited.

Recently, the emerging Large Language Models~(LLMs)~\cite{touvron2023llama, openai2023gpt4} have shown great potential for general intelligence in the general domain. Considering the great task generalization capabilities and user interactivity of LLMS, it is also expected that LLMs can give important assistance to the research of chemistry. Both \cite{zhao2024chemdfm} and \cite{zhang2024chemllm} explore the methods to incorporate LLMs into the field of chemistry. Specifically, \cite{zhao2024chemdfm} constructs a 34B-token domain-pretraining dataset and an instruction-tuning dataset containing 2.7M instructions. Based on these data, they specialize LLaMa-13B model~\cite{touvron2023llama} and get ChemDFM, one of the first chemical LLMs. ChemDFM achieves promising performances across a wide range of chemical tasks and demonstrates strong free-form dialogue capability. In addition, \cite{zhang2024chemllm} proposes a two-stage instruction-tuning pipeline and constructs a chemical LLM called ChemLLM. These chemical LLMs significantly broaden the application scope of chemical models. However, current chemical LLMs can only process pure text and SMILES notations which are also in the form of text. Considering that chemistry is intrinsically a multi-modal subject, the capabilities achievable by text-only LLMs are limited. Most recently, \cite{li2024molm} proposes 3D-MoLM which is the first generalist LMM in the field of chemistry. However, their works are still in the early stages with relatively low performances and limited modalities.

In addition, Large Multimodal Models~(LMMs) in the general domain have been rapidly developing. The advanced LMMs~\cite{mplugdocowl, internlmxcomposer,llava,gpt-4v,qwenvl,gemini} have achieved remarkable performance on a wide series of vision-language tasks. The promising capabilities of LMMs also enable their applications in more professional vertical domains. Taking the medical domain as an example, \cite{tu2023generalist} propose Med-PaLM M, the first LMMs specialized in the medical domain, and show promising performance both in quantitative evaluation and qualitative evaluation. Additionally, \cite{yang2024advancing} trained one model for each kind of modality respectively in the medical domain. However, as the models for each modality are separate, they cannot facilitate communication and collaboration between modalities.

In this work, we construct a cross-modal chemical LLM called ChemDFM-X that can comprehend and infer chemical data of multiple modalities with the same sets of parameters while handling a wide variety of tasks.




%% file: 3.chemdfm-x.tex
\section{ChemDFM-X}

In this section, we detail the methods to develop the ChemDFM-X model. We first give an overview of our training procedure~(Section~\ref{sec:overview}), introducing what modalities are involved in ChemDFM-X training and how ChemDFM-X is trained. Then, we demonstrate the training details of each modality~(Section~\ref{sec:2d}\textasciitilde\ref{sec:ir}), including data construction and modality encoder selection.

\subsection{Overview}\label{sec:overview}

\begin{figure}
    \centering
    \includegraphics[width=\textwidth,trim=50 60 50 60,clip]{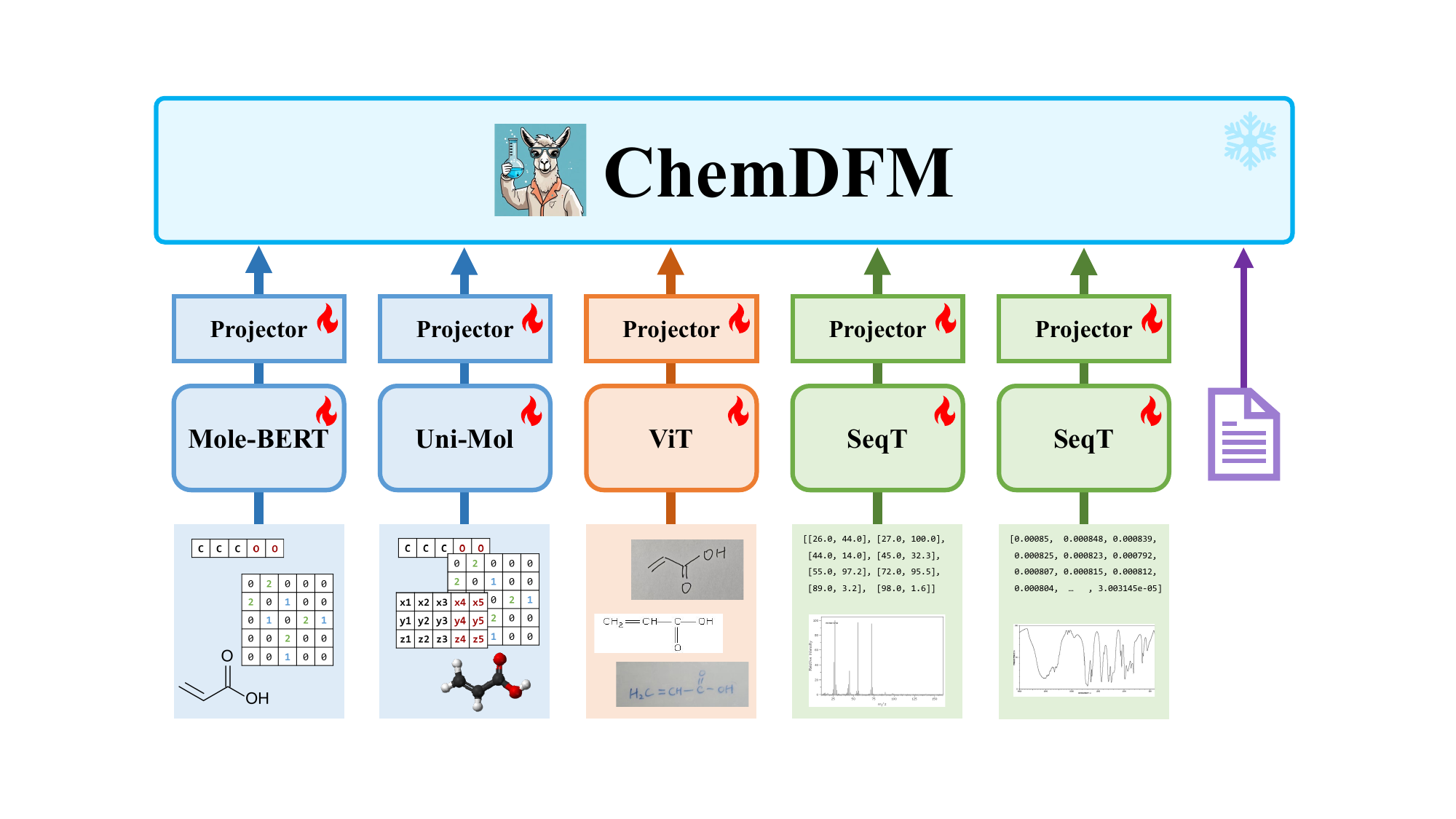}
    \caption{Overview of ChemDFM-X model structure and training paradigm. The colors to mark different input modalities are aligned with Figure 1 in the main text.}
    \label{figure:model}
\end{figure}

The overview of the structure and training paradigm of ChemDFM-X is shown in Figure~\ref{figure:model}. Generally speaking, our ChemDFM-X incorporates the typical ``LLM decoder + modality encoder'' framework widely used by current LMMs~\cite{liu2024improved,qwenvl}. Considering the significant differences in data formats among the different modalities, we incorporate separate modality encoders and corresponding projection modules for each modality. Through this ``separate encoders + unified decoder'' design, the separate encoders enable ChemDFM-X to obtain knowledge and information from different modalities, while the LLM decoder provides the capabilities to aggregate and analyze information from different modalities within a simple generative framework.


\paragraph{Modality Selection.} In the field of chemistry, there are primarily two kinds of modalities: \emph{structural} modalities and \emph{characterization} modalities.

Structural modalities mainly directly represent the connections and\slash or spatial arrangement of molecules and are usually used for reaction inference or theoretical calculation. Among the structural modalities, two modalities are introduced to ChemDFM-X, namely two-dimensional molecular graph and three-dimensional molecular conformation.

On the other hand, characterization modalities mainly imply the partial properties and substructure information of molecules. As the characterization results of molecules, their data is usually in the form of data point sequences with information implicitly hidden among them. One typical usage of characterization modality data is to identify unknown substances. There are many kinds of characterization methods in the field of chemistry. Due to the high expense of chemical experiments, the amount of real experimental data of these modalities is very limited, which significantly hinders the development of AI models. In this work, we manage to construct a great amount of tandem mass spectrum~(MS2) data and infrared spectrum~(IR) data through approximate calculation and model prediction, two of the most widely used characterization methods. ChemDFM-X is then trained on these two characterization modalities.

Apart from these four modalities, we also introduce the image modality, including molecular images and reaction images, to ChemDFM-X, as images are the most convenient data used by human researchers. Please refer to Figure~\ref{figure:overview} for examples of these modalities.

\paragraph{Model Framework.} For the LLM decoder, we use one of the advanced chemical LLMs, namely ChemDFM, to leverage the promising chemical language and notation comprehension capability it has acquired. For the modality encoders, according to the current research status of specialist models for different modalities, we either select the advanced existing specialist models or retrain one ourselves as the modality encoders. To better align the output space of modality encoders with the input space of ChemDFM, we incorporate a separate projection module for each modality. Specifically, each of the projection modules is composed of two simple linear layers with a \texttt{gelu} activation layer between them.

\paragraph{Training.} During the training process of each modal, we freeze the parameters of the pre-trained ChemDFM to maintain the advanced natural language and SMILES processing capabilities it has already acquired. Both the modality encoders and the projection modules are trained to get a better alignment between the encoder outputs and ChemDFM inputs.

The key hyperparameters we chose for each of the modalities are shown in Table~\ref{tab:hyperparameter}. The training details of each non-text modality will be introduced in the following subsections, including the instruction-tuning dataset construction process and modality encoder for each modality.

\begin{table}
    \small
    \centering
    \begin{tabular}{c|ccccc}
    \toprule
        Modality & Graph & Conformation & Image & MS2 & IR \\
    \midrule
        Modality Encoder & Mole-BERT\cite{xia2023molebert} & Uni-Mol~\cite{zhou2023unimol} & CLIP~\cite{radford2021learning} & Transformers & Transformers\\
        Projector & 2-layer MLP & 2-layer MLP& H-Reducer & 2-layer MLP& 2-layer MLP\\
        Encoding Dims & 300 & 512 & 1024 & 768 & 768 \\
        \multirow{2}{*}{\# Modality Tokens} & \# of Atoms & \# of Atoms & 72 per sub-image & \# of Peaks & 50 \\
        & (Dynamic)& (Dynamic) & (Dynamic) & (Dynamic)& (Static) \\
        Peak Learning Rate & 1e-5 & 2e-4 & 2e-3 & 2e-4 & 2e-4 \\
        Total Batch Size & \multicolumn{5}{c}{256} \\
        Training Epochs & \multicolumn{5}{c}{3} \\
    \bottomrule
    \end{tabular}
    \caption{Overview of the key hyperparameters of ChemDFM-X.}
    \label{tab:hyperparameter}
\end{table}

\subsection{Structural Modality: Molecular Graph}\label{sec:2d}

The molecular graph is a typical structural modality for molecules. It only consists of the type of atoms in the molecule and the connections between them. Considering that the SMILES notation also contains these two kinds of information, the molecular graph can be easily obtained from the SMILES notation of the corresponding molecule according to simple rules. Formally, a molecular graph can be defined as an undirected graph $\mathcal{G} = (\mathbf{V}, \mathbf{E})$, where $\mathbf{V} = (v_1, v_2, \dots, v_{|\mathbf{V}|})$ is the atom set with $v_i$ representing the element type of the $i$-th atom and $\mathbf{E} = (\boldsymbol{e}_1, \boldsymbol{e}_2, \dots, \boldsymbol{e}_{|\mathbf{E}|})$ is the chemical bond set with $\boldsymbol{e}_i=(v_m, v_n, b_i)$ indicating that a chemical bond of type $b_i$ exists between atoms $v_m$ and $v_n$.

\paragraph{Data Construction.}
Considering that SMILES notation can also be viewed as a special structural modality and the molecule graphs can be easily obtained from SMILES, the SMILES-related instruction-tuning dataset used by traditional Chemical LLM is naturally suitable for the training of the molecular graph modality. Therefore, we follow the dataset composition and collection procedure proposed by ChemDFM~\cite{zhao2024chemdfm} to construct the instruction-tuning data for the molecular graph modality.

Specifically, the data we construct for the instruction tuning of molecular graph modality is mainly composed of the following:
\begin{itemize}
    \item \textbf{Molecule description.} In this task, models are asked to describe the molecule based on the given molecular representation. The molecule-description pairs are collected from PubChem\footnote{\url{https://pubchem.ncbi.nlm.nih.gov/}}, a web-scale chemical database that contains more than 100M compounds. For each molecule-description pair, we construct two sets of data: one using the molecular graph alone as the molecular representation, and the other using both the molecular graph and the SMILES notation. To further increase the quality of this data, the high-quality molecule-description pairs (where the descriptions have more than two sentences) are repeated twice.
    \item \textbf{Molecular property prediction.} In these tasks, models are asked to predict the properties of the given molecule based on its representation. The data are constructed from the most widely used molecular property prediction benchmark, Molecule Net~\cite{moleculenet}. Similar to the molecule description dataset, we generate the molecular graphs based on the provided SMILES and construct two sets of data using the molecular graph alone or with SMILES.
    \item \textbf{Reaction completion.} One of the most important usages of molecular graphs is reaction representation and inference. In this task, models are asked to complete an incomplete reaction where part of the involved molecules are missing. The source reactions are sampled from USPTO~\cite{uspto}, the largest chemical reaction database. For each reaction, we randomly mask the reactant, reagent, or product, where the remaining molecules of the reaction are represented by molecular graph alone or with SMILES.
    \item \textbf{Molecular graph recognition.} In these tasks, models are asked to recognize the molecule graph by providing the alternative representation of the corresponding molecule, such as SMILES, IUPAC\footnote{short for International Union of Pure and Applied Chemistry} name, or molecular formula. The involved molecules are sampled from PubChem. Based on the molecules, we construct five sub-tasks: predicting the IUPAC name given the molecular graph alone or with SMILES, predicting the molecular formula given the molecular graph alone or with SMILES, and predicting the SMILES notation given the molecular graph alone.
\end{itemize}

\begin{table}
    \centering
    \begin{tabular}{ccccccc}
    \toprule
        \multirow{2}{*}{Data Type} & \multirow{2}{*}{\# Samples} & \multirow{2}{*}{Data Source} & \multicolumn{4}{c}{\# Heavy Atoms in each Molecule} \\
        & & & Max & Min & Average & Median \\
    \midrule
        Molecule Description & 1152K & PubChem & 574 & 1 & 35.0 & 29 \\
        Molecular Property Prediction & 203K & MoluculeNet & 222 & 1 & 24.0 & 22\\
        Reaction Completion & 600K & USPTO & 273 & 1 & 23.3 & 22 \\
        Molecular Graph Recognition & 30K & PubChem & 150 & 1 & 19.5 & 18\\
    \bottomrule
    \end{tabular}
    \caption{Composition of the instruction-tuning dataset for Molecular Graph and the statistics of the molecules involved}
    \label{tab:data2d}
\end{table}

\begin{figure}
    \centering
    \includegraphics[width=0.6\textwidth,trim=100 15 100 15,clip]{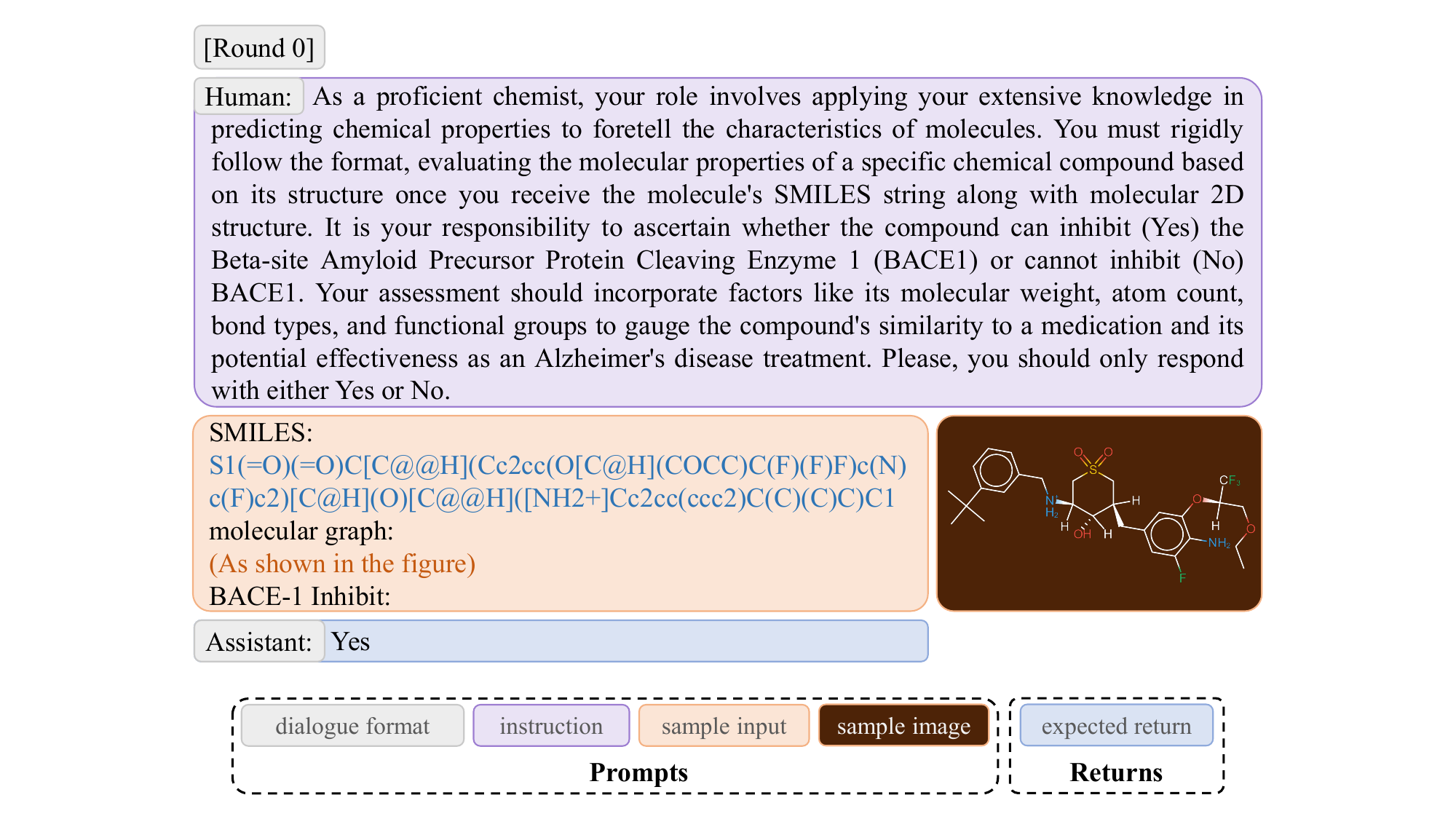}
    \caption{An example of the final structure of instruction tuning data.}
    \label{figure:sample1}
\end{figure}

It is worth noticing that we carefully remove all the data that may be present in the evaluations based on SMILES matching. For the prompt, we leverage the same dialogue format used by the ChemDFM, while rewriting the instruction of each task using GPT-4 to diversify the expressions.
The final composition and statistics of the instruction tuning dataset are demonstrated in Table~\ref{tab:data2d} with a detailed example illustrated in Figure~\ref{figure:sample1}.

\paragraph{Graph Encoder.} We utilize one of the most recent and advanced models, namely Mole-BERT~\cite{xia2023molebert}, which specializes in 2D molecular graph understanding. The model is an encoder-decoder based graphic neural network (GNN) pre-trained on 2 million molecules
and achieves promising performances on multiple downstream tasks.
Specifically, the Mole-BERT model follows an encoder-decoder architecture where the encoder and the decoder each consist of 5-layer graph isomorphism networks (GINs). We adopt its encoder as our molecular graph encoder and utilize the resulting node representations for each atom as the feature of the molecular graphs.

\subsection{Structural Modality: Molecular Conformation}\label{sec:3d}

\begin{table}[t]
    \centering
    \begin{tabular}{ccccccc}
    \toprule
        \multirow{2}{*}{Data Type} & \multirow{2}{*}{\# Samples} & \multirow{2}{*}{Data Source} & \multicolumn{4}{c}{\# Heavy Atoms in each Molecule} \\
        & & & Max & Min & Average & Median \\
    \midrule
        Molecule Description & 1152K & PubChem & 574 & 1 & 35.0 & 29 \\
        Molecular Property Prediction & 203K & MoluculeNet & 222 & 1 & 24.0 & 22\\
        Reaction Completion & 600K & USPTO & 273 & 1 & 23.3 & 22\\
        Molecular Conformation Recognition & 30K & PubChem & 150 & 1 & 19.5 & 18\\
    \bottomrule
    \end{tabular}
    \caption{Composition of the instruction-tuning dataset for Molecular Conformation and the statistics of the molecules involved}
    \label{tab:data3d}
\end{table}

The molecular conformation is another typical structural modality for molecules. Compared with molecular graphs, molecular conformations additionally contain the coordinate information of each atom of the molecules, reflecting the position information of atoms in the molecule. Therefore, molecular conformations usually contain more information than molecular graphs and SMILES notations. Formally, a molecular conformation can be viewed as a similar undirected graph to molecular graphs $\mathcal{G} = (\mathbf{V}^\prime, \mathbf{E})$ with $\mathbf{V}^\prime = (\boldsymbol{v}_1, \boldsymbol{v}_2, \dots, \boldsymbol{v}_{|\mathbf{V}|})$ and $\boldsymbol{v}_i=(x_i, y_i, z_i, a_i)$ where $(x_i, y_i, z_i)$ is the Cartesian coordinates of the $i$-th atom and $a_i$ denotes the element type of the $i$-th atom.


\paragraph{Data Construction.} We construct the instruction-tuning dataset for molecular conformation modality based on the dataset used for molecular graph modality. Specifically, we utilize the same source data as molecular graphs while replacing all the molecular graphs in the inputs with molecular conformations. We also use the same input format while instruction is rewritten specifically for molecular conformations.
The final composition and statistics of the instruction tuning dataset are demonstrated in Table~\ref{tab:data3d}

However, it is worth noticing that obtaining molecular conformations from SMILES is a lot more complicated than obtaining molecular graphs from SMILES. Ideally, rigorous calculations based on quantum chemistry are needed to acquire the ``optimal'' conformations where the potential energy of the molecule is the lowest. However, quantum chemistry calculation can be very costly, especially for molecules with a large amount of atoms. To tackle this problem, we follow the same approximate calculation process used by Uni-Mol~\cite{zhou2023unimol} to obtain the approximated conformations under optimization algorithms. Specifically, we first generate the original molecular conformations using RDKit~\cite{rdkit}. Then, the molecular conformation is optimized using the Merck Molecular Force Field (MMFF)~\cite{halgren1996merck} algorithm, and the pseudo-optimal conformation for the molecule is obtained.

\paragraph{Conformation Encoder.}
We adopt one of the famous large-scale molecular conformation representation models, Uni-Mol~\cite{zhou2023unimol}, as our 3D molecular conformation encoder. It is a transformer-based 3D molecular pre-trained model derived from 19M molecules and 209M molecular conformations. It achieved promising results in various downstream tasks, especially in 3D spatial tasks.

\subsection{Image Modality}\label{sec:fig}

During the work of human chemists, images, rather than structured molecular graphs and conformations, are the most used modality through which they perceive and express molecules and reactions. Therefore, it is vital for the chemical LLM to possess image comprehension capability so that the chemist can communicate with them more conveniently.

\paragraph{Data Construction.} We utilize the same data sources as molecular graphs and molecular conformations to construct the instruction-tuning dataset for the image modality by replacing the non-text molecular representation with images and generating corresponding instructions.

\begin{figure}[!t]
	\centering
	\subfloat[RDKit Style Image]{
		\includegraphics[width=0.3\linewidth,trim=0 0 0 0,clip]{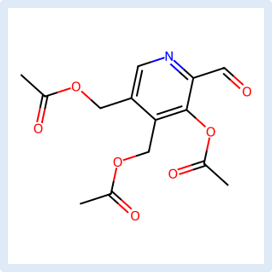}}
	\subfloat[Indigo Style Image]{
		\includegraphics[width=0.3\linewidth,trim=0 0 0 0,clip]{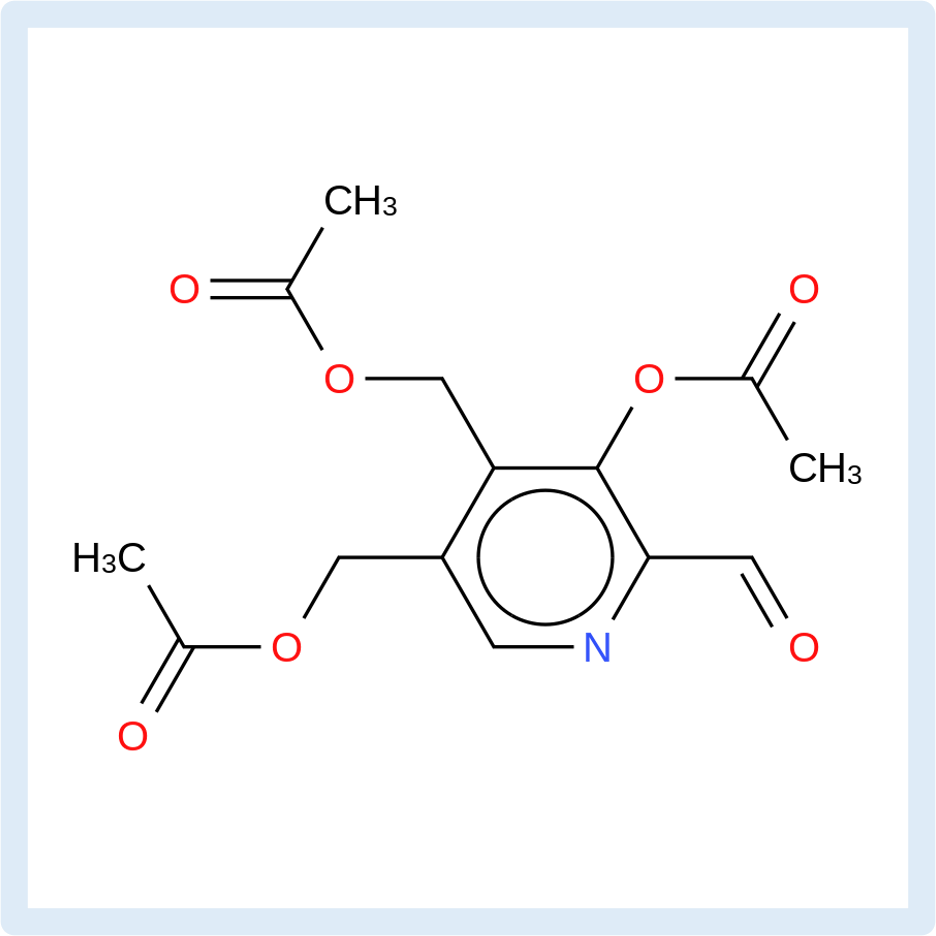}}
	\subfloat[Handwritten Style Image]{
		\includegraphics[width=0.3\linewidth,trim=0 0 0 0,clip]{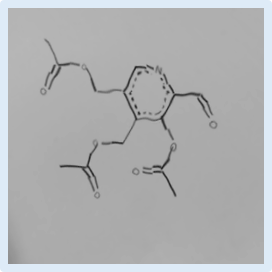}}
	\caption{Examples of three molecular styles.}
	\label{figure:styles}
\end{figure}

Moreover, to diversify the styles of images, we employed three different methods for the generation of molecular images. Firstly, we directly use two different toolkits namely RDKit~\cite{rdkit} and Indigo~\cite{indigo}. We further conduct traditional image augmentation methods, such as random grayscale conversion and random color jitter, on the resulting images. Besides, to familiarize ChemDFM-X with the handwritten style images, we utilize the pipeline proposed in \cite{chempix} to generate pseudo-handwritten images. The comparison of the three styles of images is illustrated in Figure~\ref{figure:styles}.

In addition to the existing tasks that revolve around molecule images, we add a new task focusing on the reaction images. Specifically, we randomly sample 300K reactions from the USPTO database and generate the images of the reactions using the RDKit toolkit. Based on the reaction image alone, the model is asked to recognize the reaction by providing the SMILES notation of the whole reaction.

The final composition and statistics of the instruction tuning dataset are demonstrated in Table~\ref{tab:dataim}.

\begin{table}
    \centering
    \begin{tabular}{ccccccc}
    \toprule
        \multirow{2}{*}{Data Type} & \multirow{2}{*}{\# Samples} & \multirow{2}{*}{Data Source} & \multicolumn{4}{c}{\# Heavy Atoms in each Molecule} \\
        & & & Max & Min & Average & Median \\
    \midrule
        Molecule Description & 1152K & PubChem & 574 & 1 & 35.0 & 29 \\
        Molecular Property Prediction & 204K & MoluculeNet & 222 & 1 & 24.0 & 22 \\
        Reaction Completion & 600K & USPTO & 273 & 1 & 23.3 & 22 \\
        Molecular Image Recognition & 30K & PubChem & 150 & 1 & 19.5 & 18 \\
        Reaction Image Recognition & 300K & USPTO & 440 & 1 & 23.7 & 22 \\
    \bottomrule
    \end{tabular}
    \caption{Composition of the instruction-tuning dataset for Image modality and the statistics of the molecules involved}
    \label{tab:dataim}
\end{table}

\paragraph{Image encoder.} There is no previous work that specifically targeted both molecule images and reaction images. Therefore, we employ CLIP~\cite{radford2021learning}, the image encoder widely used in general domain LMMs. The original CLIP model can only process images with resolutions of $224\times224$ or $336\times336$. It is far from being sufficient for chemical images, especially for images of large molecules and reactions. Pre-experiments show that forcibly compressing images to these resolutions results in the model learning almost nothing. To handle this problem, we follow the method proposed by \cite{llava}. Specifically, instead of directly resizing the images of original resolution, we crop them into multiple sub-images that fit the input size of CLIP and then leverage a specific module named H-Reducer~\cite{hu2024mplug} to reduce the output sequence length of CLIP by a factor of $n$. In our cases, $n$ is set to $8$.

\subsection{Characterization Modality: MS2 Spectrum}\label{sec:ms}

\begin{table}[t]
    \centering
    \begin{tabular}{ccccccc}
    \toprule
        \multirow{2}{*}{Data Type} & \multirow{2}{*}{\# Samples} & \multirow{2}{*}{Data Source} & \multicolumn{4}{c}{\# Heavy Atoms in each Molecule} \\
        & & & Max & Min & Average & Median \\
    \midrule
        MS2 spectrum recognition & 21K & PubChem & 99 & 1 & 18.3 & 18\\
        Molecular Property Prediction & 90K & MoluculeNet & 135 & 1 & 23.4 & 22 \\
        Reaction Related MS2 Identification &  224K & USPTO & 131 & 1 & 23.5 & 22\\
    \bottomrule
    \end{tabular}
    \caption{Composition of the instruction-tuning dataset for MS2 Spectrum and the statistics of the molecules involved}
    \label{tab:datams}
\end{table}

Molecular graphs, molecular conformations, and molecular images provide information about the structure of known molecules, however, numerous molecules of unknown structure exist in nature or may appear during experiments. In this context, spectroscopy becomes one of the key tools to determine the identity of a molecule. Mass spectrum is one of the common types of spectra, which is based on the principle that the molecule is ionized by an ion source and then cleaved into ion fragments with different mass-to-charge ratios.\footnote{\url{https://en.wikipedia.org/wiki/Mass_spectrum}} The mass spectrum consists of a series of ion peaks, where the horizontal axis represents the mass-to-charge ratio of the ion fragments and the vertical axis represents the intensity of the peaks. Formally, the MS2 spectrum data can be represented by a tuple sequence $\mathbf{M} = ((r_1, I_1), (r_2, I_2), \dots, (r_n, I_n))$, where $(r_i, I_i)$ denotes the $i$-th peak, $r_i$ denotes the mass-to-charge ratio of the $i$-th peak, $I_i$ denotes the intensity of the $i$-th peak, and $n$ represents the total number of peaks in the MS2 spectrum. By introducing mass spectrum data, the ability to deal with unknown molecules can be added to the chemical LLM.

\paragraph{Data Construction.} As a typical kind of characterization modalities, the tasks suitable for MS2 spectra differ from those for the structural modalities. Therefore, to construct the instruction-tuning dataset for MS2 spectra, we select a small subset of tasks that are suitable for MS2 spectra from our instruction-tuning dataset for structural modalities, and supplement them with tasks specific to characterization modalities. Specifically, the instruction-tuning dataset for MS2 spectra is composed of the following tasks:
\begin{itemize}
    \item \textbf{MS2 spectrum recognition.} We utilize the same data source as the molecular graph recognition task introduced in Section~\ref{sec:2d} with the molecular graph replaced by the MS2 spectra of the corresponding molecule.
    \item \textbf{Molecular property prediction.} The constructing process is also akin to that of the molecular graph modality. However, different from molecular graphs, it does not make sense to predict the molecular properties using MS2 spectra alone. Therefore, we only instruct the model to conduct the prediction based on the MS2 spectra along with the SMILES notations.
    \item \textbf{Reaction related MS2 spectrum identification.} In practice, one of the most important usages of the spectra is to identify unknown molecules in the context of a reaction. 
    In such scenarios, the identity of the unknown molecules should be inferred in accordance with not only the spectra, but also the a priori information of the reaction, like the known reactants or the detected products.
    Therefore, the capability to identify the MS2 spectrum with the help of the reaction related to the molecule is also important and valuable for chemical LMMs. To construct the data for such tasks, we reuse the data for reaction completion tasks in the instruction-tuning datasets of structural modalities. The incomplete reactions will be represented by SMILES notations, while the MS2 spectra of the missing substances are also provided to the model. Based on this information, the model is instructed to identify the corresponding molecules by providing the SMILES of the molecules.
\end{itemize}

Another great challenge for the MS2-spectrum modality is the great expense associated with obtaining real experimental data due to the significant human labor required throughout the experimental process. Therefore, large-scale experimental MS2 spectra data is absent. To tackle this problem, we use a prediction-based approach to generate mass spectrum data in large batches. The prediction tool we used is CFM-ID 4.0~\cite{cfmid4.0}, which provides a method for accurate and efficient prediction of molecular MS2 spectra by creating a probabilistic graphical generative model for the MS2 fragmentation process through competitive fragmentation modeling, and adjusting the model parameters from the data through machine learning algorithms. With CFM-ID 4.0, we obtain a total of about 700K tandem mass spectra with their corresponding SMILES to form the instruction tuning dataset for for training the encoder of MS2 spectrum modality. The final composition and statistics of the instruction tuning dataset are demonstrated in Table~\ref{tab:datams}

\paragraph{MS2 Sequence Transformer.}

Since there is little work on encoding MS2 spectra, we choose to train a new spectrum encoder for MS2. Considering the data format of MS2 spectra is point sequences and one of the classical applications of the spectrum is structural inference~\cite{Demartini13}, we train an encoder-decoder manner transformer model with the pertaining task designed as a spectral inference task. Specifically, the model is trained to generate the SMILES notation of the molecule corresponding to the input MS2 spectrum. We build the codebook of the tokens based on the transverse coordinates and then acquire the input token list from the ion peaks of the MS2 spectrum using the codebook. After the model is trained, we use its encoder module as the encoder for the MS2 spectrum modality.

\subsection{Characterization Modality: IR Spectrum}\label{sec:ir}

Similar to the MS2 spectrum, the IR spectrum also contains rich chemical information about the molecule and is one of the most important means to characterize the molecular structure. The principle of the infrared spectrum is that a molecule can selectively absorb infrared rays of certain wavelengths, causing its own vibrational and rotational energy levels transitions, which is related to the detailed structure of molecules\footnote{\url{https://en.wikipedia.org/wiki/Infrared_spectroscopy}}. The infrared spectra of molecules can be obtained through the detection of infrared absorption. The horizontal coordinate of the infrared spectrum indicates the wavelength of the infrared ray, and the vertical coordinate generally implies the absorption intensity. Formally, the IR spectrum can be represented by a point sequence $\mathbf{R} = ((w_1, t_1), (w_2, t_2), \dots, (w_l, t_l))$, where $w_i$ denotes the wave length of the light, $t_i$ is the absorption intensity of the corresponding light.

\begin{table}
    \centering
    \begin{tabular}{ccccccc}
    \toprule
        \multirow{2}{*}{Data Type} & \multirow{2}{*}{\# Samples} & \multirow{2}{*}{Data Source} & \multicolumn{4}{c}{\# Heavy Atoms in each Molecule} \\
        & & & Max & Min & Average & Median \\
    \midrule
        IR spectrum recognition & 30K & PubChem & 150 & 1 & 19.5 & 18\\
        Molecular Property Prediction & 102K & MoluculeNet & 222 & 1 & 24.0 & 22\\
        Reaction Related IR Identification & 300K & USPTO & 440 & 1 & 22.9 & 22\\
    \bottomrule
    \end{tabular}
    \caption{Composition of the instruction-tuning dataset for IR Spectrum and the statistics of the molecules involved}
    \label{tab:datair}
\end{table}

\paragraph{Data Construction.} We construct the instruction-tuning dataset for the IR-spectrum modality based on the dataset used for the MS2-spectrum modality. Specifically, we utilize the same source data while replacing all the MS2 spectra with IR spectra with corresponding modifications of the instructions.

Similar to MS2 spectra, IR spectra also suffer from the high cost of the experimental data. To make things even worse, the expense of IR spectra calculation is also beyond acceptable. Because the simulation of molecular infrared spectra needs to go through the time-consuming DFT process and the large-volume infrared computation puts higher demands on the arithmetic power. Based on the above considerations, we adopt a neural network prediction-based approach, Chemprop-IR~\cite{Chemprop-IR}, to generate IR data. We generate about 1M IR spectra in total to form the instruction-tuning dataset for training the encoder of IR spectrum modality. The final composition and statistics of the instruction tuning dataset are demonstrated in Table~\ref{tab:datair}

\paragraph{IR Sequence Transformer.}

Similar to the situation of MS2 spectra, there is little work associated with the encoding of IR spectra. Therefore, we choose to train a new spectrum encoder for IR. We used the same setup as the mass spectrum including the model structure and the pertaining task. The data in the IR spectrum are generally denser than MS2 spectra, as the MS2 spectra data are composed of only the peaks while the IR spectra are composed of all the data points. Therefore, instead of the codebook tokenization strategy used for MS2 spectra, we directly construct the input features by reshaping the sequence of absorption intensity. After the model is trained, we use its encoder module as the encoder for the IR spectrum modality.


%% file: 4.evaluation.tex
\section{Evaluation}

Our evaluations are mainly conducted based on ChemLLMBench~\cite{guo2023what}, one of the most widely used benchmarks designed for the evaluation of LLMs in the field of chemistry. However, considering the intrinsic differences across the modalities, the types of tasks suitable for each modality may differ. For example, the information contained in characterization modalities is often more obscure and scattered compared to that in structural modalities, therefore, it makes less sense to ask models to generate molecular captions based on the molecular spectra. Therefore, to thoroughly and specifically evaluate ChemDFM-X's capabilities across different modalities, we construct separate evaluation task sets for each modality by both selecting existing tasks from ChemLLMBench and building new tasks. In this section, we will introduce the evaluation tasks and analyze model performances for each kind of modality separately.

\subsection{Structural Modalities}

\begin{figure}
    \centering
    \includegraphics[width=\linewidth]{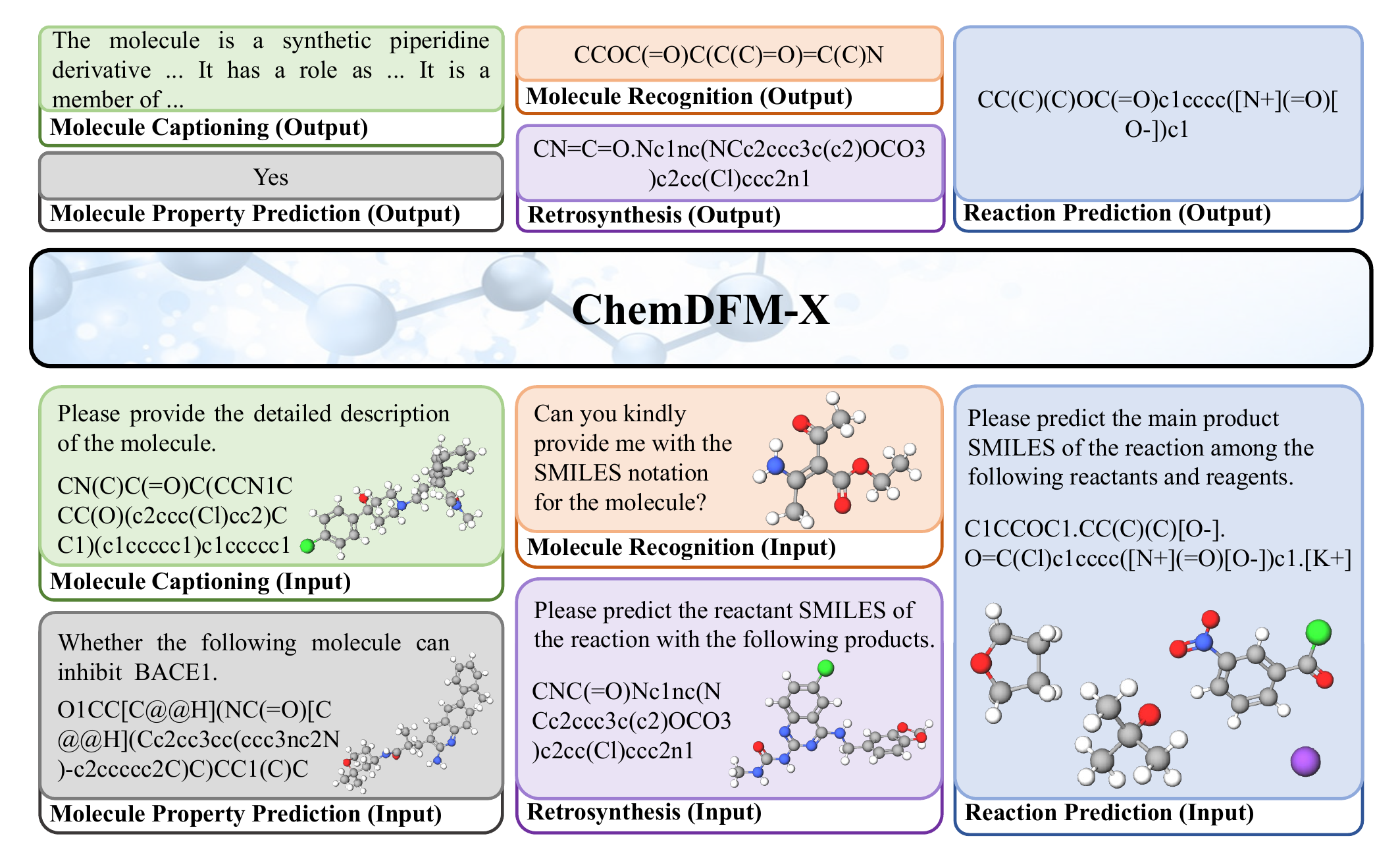}
    \caption{Evaluation tasks for structural modalities}
    \label{figure:eval1}
\end{figure}

\begin{table*}
    \centering
    \setlength\tabcolsep{5pt}
    \begin{tabular}{lcccccccc}
    \toprule
    \multirow{2.1}{*}{Model} & $\text{MR}^*$ & \multicolumn{3}{c}{Molecule Captioning} & \multicolumn{2}{c}{RP} & \multicolumn{2}{c}{Retrosynthesis} \\
    \cmidrule(rl){2-2} \cmidrule(rl){3-5} \cmidrule(rl){6-7} \cmidrule(rl){8-9}
    & \small{Acc} & \small{BLEU-2} & \small{BLEU-4} & \small{METEOR} & \small{Acc} & \small{Validity} & \small{Acc} & \small{Validity} \\
    \midrule
    \rowcolor{grey}\multicolumn{9}{c}{\textit{specialist models}} \\
    Text+Chem T5 (S)~\cite{christofidellis2023unifying} & - & 62.5 & 54.2 & 64.8 & - & - & - & - \\
    MolCA (S + G)~\cite{liu2023molca} & - & 63.9 & 55.5 & 66.9 & - & - & - & - \\
    Chemformer (S)~\cite{chemformer} & - & - & - & - & 93.8 & 100 & 53.6 & 100 \\
    \midrule
    \rowcolor{grey}\multicolumn{9}{c}{\textit{generalist models}} \\
    ChemLLM-7B-Chat (S)~\cite{zhang2024chemllm} & - & 14.5 & 7.4 & 22.2 & 18.0 & 94.0 & 10.0 & 91.0 \\
    ChemDFM-v1.0 (S)~\cite{zhao2024chemdfm} & - & 26.1 & 20.6 & 33.1 & 49.0 & 98.0 & 12.0 & \textbf{99.0} \\
    3D-MoLM (S + C)~\cite{li2024molm} & 0 & 18.2 & 9.9 & 25.6 & $\text{-}^\dag$ & - & - & -  \\
    \multirow{1}{*}{\textbf{ChemDFM-X (G)}} & 41.0 & 19.3 & 13.2 & 26.2 & 29.0 & 94.0 & 9.0 & 94.0 \\
    \multirow{1}{*}{\textbf{ChemDFM-X (S + G)}} & - & \underline{26.6} & \underline{21.1} & \underline{33.8} & 48.0 & \textbf{99.0} & \textbf{18.0} & 98.0 \\
    \multirow{1}{*}{\textbf{ChemDFM-X (C)}} & 66.0 & 26.4 & 20.7 & 33.6 & \underline{53.0} & \textbf{99.0} & 8.0 & 97.0 \\
    \multirow{1}{*}{\textbf{ChemDFM-X (S + C)}} & - & \textbf{27.8} & \textbf{22.0} & \textbf{34.7} & \textbf{54.0} & \textbf{99.0} & \underline{16.0} & \textbf{99.0} \\
    \bottomrule
    \end{tabular}
    \caption{The Results of multiple evaluation tasks, including Molecule Recognition~(MR), Molecule Captioning, Reaction Prediction~(RP), and Retrosynthesis, for the structural modalities. The metric validity evaluates whether the output SMILES is valid. The content within the parentheses indicates the molecular representation modalities used for the corresponding model (S - SMILES, G - Molecular Graphs, C - Molecular Conformations). *: The SMILES notations are not included in the inputs of this task to prevent cheating. $\dag$: 3D-MoLM does not support the reaction-related tasks as the input of 3D-MolM can not contain more than one molecular conformation. Among the generalist models, the optimal results are \textbf{bolded}, and the second-best results are \underline{underlined}.}
    \label{tab:strcut-other}
\end{table*}
\begin{table*}
    \centering
    \begin{tabular}{lcccccc}
    \toprule
    Model & BACE & BBBP & Clintox & HIV & Tox21 & Avg \\
    \midrule
    \rowcolor{grey}\multicolumn{7}{c}{\textit{specialist models}} \\
    Mole-BERT (G)~\cite{xia2023molebert} & 80.8 & 71.9 & 78.9 & 78.2 & 76.8 & 77.3 \\
    Uni-Mol (C)~\cite{zhou2023unimol} & 85.7 & 72.9 & 91.9 & 80.8 & 79.6 & 82.2 \\
    MolXPT (S)~\cite{liu-etal-2023-molxpt} & 88.4 & 80.0 & 95.3 & 78.1 & 77.1 & 83.8 \\
    MolCA (S + G)~\cite{liu-etal-2023-molca} & 79.8 & 70.0 & 89.5 & 64.5 & 77.2 & 76.2 \\
    \midrule
    \rowcolor{grey}\multicolumn{7}{c}{\textit{generalist models}} \\
    ChemLLM-7B-Chat (S)~\cite{zhang2024chemllm} & \textbf{83.2} & 62.8 & 68.5 & 73.8 & 70.1 & 71.7 \\
    ChemDFM-v1.0 (S)~\cite{zhao2024chemdfm} & 78.4 & \textbf{66.7} & \underline{89.9} & 73.6 & \textbf{79.8} & \underline{77.7} \\
    3D-MoLM (S + C)~\cite{li2024molm} & 50.9 & 48.0 & 47.0 & 53.2 & 49.4 & 49.7\\
    \multirow{1}{*}{\textbf{ChemDFM-X (G)}} & 66.6 & 65.2 & 45.5 & 70.8 & 79.0 & 65.4 \\
    \multirow{1}{*}{\textbf{ChemDFM-X (S + G)}} & 76.1 & \underline{65.6} & \textbf{92.5} & \underline{75.3} & 79.3 & \textbf{77.8} \\
    \multirow{1}{*}{\textbf{ChemDFM-X (C)}} & \underline{80.5} & 65.4 & 71.9 & 75.2 & 79.1 & 74.4 \\
    \multirow{1}{*}{\textbf{ChemDFM-X (S + C)}} & 79.0 & 63.3 & 89.5 & \textbf{76.4} & \underline{79.5} & 77.5 \\
    \bottomrule
    \end{tabular}
    \caption{The Results of molecule property prediction tasks for the structural modalities. We report the performance of the models in AUC-ROC, which stands for the Area Under the Curve of the Receiver Operating Characteristic. The content within the parentheses indicates the molecular representation modalities used for the corresponding model (S - SMILES, G - Molecular Graphs, C - Molecular Conformations). Among the generalist models, the optimal results are \textbf{bolded}, and the second-best results are \underline{underlined}.}
    \label{tab:struct-molnet}
\end{table*}

\paragraph{Evaluation Tasks.} Considering that SMILES can be viewed as a special structural modality and ChemLLMBench is originally designed for evaluating LLMs' SMILES comprehension capabilities, all the tasks in ChemLLMBench should be suitable for the evaluation of molecular graph and molecular conformation modalities. Specifically, the evaluation tasks we utilize for both structural modalities are as follows:
\begin{itemize}
    \item \textbf{Molecule Recognition.}\footnote{Corresponding to the name prediction tasks in \cite{guo2023what}} There are four types of molecule recognition tasks in ChemLLMBench. However, to focus tightly on evaluating non-text modality comprehension capabilities, we only source data of IUPAC to SMILES task and construct a new task where models need to recognize the molecular graphs or conformations of the molecule and give its SMILES notation.
    \item \textbf{Molecule Captioning.} This task requires models to generate a brief description of the given molecules. In ChemLLMBench, the molecule captioning dataset is composed of 100 instances sampled from the test set of CHEBI-20~\cite{edwards-etal-2021-text2mol}. Instead of directly using this dataset provided by ChemLLMBench, we evaluate the models on the full test set of CHEBI-20 for a more robust assessment.
    \item \textbf{Molecule Property Prediction.} These tasks ask models to predict the properties of the given molecules. To construct such tasks, ChemLLMBench extracts five typical classification task sets (\textit{i.e.}, BACE, BBBP, Clintox, HIV, and Tox21) from Molecular Net~\cite{moleculenet}, the widely used property prediction benchmark, and randomly sampled 100 molecule-property pairs from each task. In our evaluation, we use the same five task sets. However, to make the tasks more difficult, instead of randomly sampling, we incorporate a more challenging dataset split using the DeepChem library~\cite{deepchem}.
    \item \textbf{Reaction Prediction and Retrosynthesis.} These two tasks are similar in form but differ significantly in difficulty. The reaction prediction task asks models to predict the product of a chemical reaction given its reactants and reagents, which is relatively straightforward and easy. On the other hand, the retrosynthesis task asks models to predict the reactants of a reaction given its product. In our evaluation, we directly utilize the corresponding dataset provided by ChemLLMBench to evaluate these capabilities.
\end{itemize}

An intuitive diagram demonstrating the input and output of the above tasks is shown in Figure~\ref{figure:eval1}.

\paragraph{Baselines.} We leverage three types of models as our baselines. 

The most relevant baseline in the structure modalities is 3D-MoLM~\cite{li2024molm}, which is a generalist modal that can process both SMILES and molecular conformations\footnote{They also trained a 2D-MoLM which can process SMILES and molecular graphs, however, the parameter of 2D-MoLM is not yet open-sourced.}. The primary distinctions between ChemDFM-X and 3D-MoLM can be summarized as follows: First, ChemDFM-X employs a non-compressive MLP projector, whereas 3D-MoLM uses a Q-Former module which usually leads to information losses and requires more training~\cite{yao2024decodecouplingtokencompression}. Second, 3D-MoLM supports only a single molecular conformation per input, preventing it from addressing tasks involving multiple molecules, such as reaction-related problems and multi-turn dialogues. In contrast, ChemDFM-X does not suffer from this intrinsic drawback as it is designed to handle the input with multiple molecular conformations in a single prompt. Furthermore, it is also worth noticing that ChemDFM-X also supports other modalities beyond molecular conformation, which 3D-MoLM does not.

The second type of baselines is other generalist models in the field of chemistry. Specifically, we use ChemDFM~\cite{zhao2024chemdfm} and ChemLLM~\cite{zhang2024chemllm}, two pioneer works in the field of LLM for chemistry. They excel at SMILES comprehension, but can not understand any other chemical modalities.

Finally, we also list the advanced performances of task-specific specialist models. They usually can achieve excellent performance on the targeted tasks while having zero performance on the others.

\paragraph{Results and Analysis.} The experimental results of the structural modalities are illustrated in Table~\ref{tab:strcut-other} and \ref{tab:struct-molnet}. The results in Table~\ref{tab:strcut-other} show that our ChemDFM-X possesses the capability to comprehend and infer over molecular graph modality and molecular conformation modality. ChemDFM-X outperforms the baseline generalist LMM, 3D-MoLM, in all the settings while achieving better or comparable performance compared with the state-of-the-art~(SOTA) performances of generalist models. Specifically, ChemDFM-X achieves the new SOTA method among the generalist models when providing both SMILES notations and molecular conformations.

However, it is worth noticing that ChemDFM-X performs relatively better with molecular conformations compared with molecular graphs. On the one hand, from a chemical perspective, molecular conformations intrinsically contain more information than molecular graphs. Therefore, the potential of molecular conformation modality as a molecular representation is greater. On the other hand, ChemDFM-X (G) only achieves 14\% accuracy in the molecular recognition tasks, showing that the molecular graph modality may still remain undertrained in ChemDFM-X.

Another promising result is that the performances of ChemDFM-X in reaction-related tasks have improved significantly compared with those with only SMILES inputs. This result is consistent with the chemical intuition that molecular graphs and conformations are suitable and commonly used for chemical reaction inference. 

As for the molecular property prediction tasks (Table~\ref{tab:struct-molnet}), the molecular graph modality and the molecular conformation modality do not have a remarkable influence on the performances.
It is worth noticing that all the involved tasks are about high-level biological properties, such as blood-brain barrier penetration and toxicity, and there exists a remarkable overlap in the information between structural modalities including SMILES. Therefore, we attribute the reason to that the molecular graphs and confirmations fail to provide sufficient complementary information to enhance the molecule modeling for these tasks. Thus, to further increase the performance of these tasks, more comprehensive training or more powerful molecular representations are needed.

\subsection{Image Modality}

\begin{figure}
    \centering
    \includegraphics[width=0.75\linewidth]{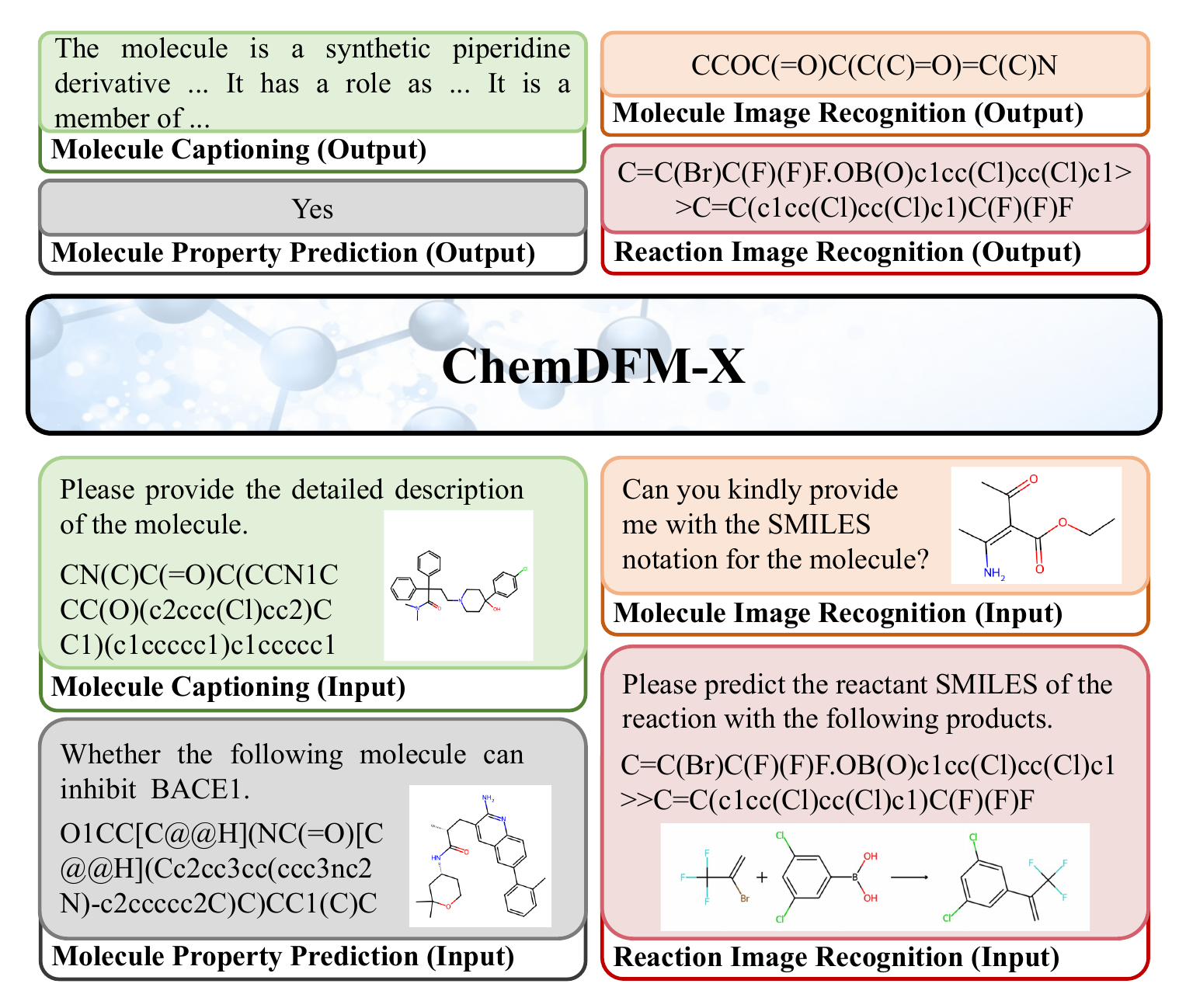}
    \caption{Evaluation tasks for image modality}
    \label{figure:eval2}
\end{figure}

\paragraph{Evaluation Tasks.} Considering that image modality is the most convenient modality for human researchers, our evaluations of image modality focus more on whether models can recognize the given image. Specifically, we evaluate both the molecule image recognition capability and the reaction image recognition capability of the models. The molecule image recognition task is intuitively constructed from the molecule recognition dataset used by structural modalities. As for the reaction image recognition task, we randomly sample 100 reactions from the USPTO-50K dataset~\cite{uspto} to form its test set. Besides, we also evaluate the molecule captioning and molecule property prediction tasks with the image modality inputs. An intuitive diagram demonstrating the input and output of the evaluation tasks for image modality is shown in Figure~\ref{figure:eval2}.

\begin{table*}
    \centering
    \setlength\tabcolsep{5pt}
    \begin{tabular}{lccccc}
    \toprule
    \multirow{2.1}{*}{Model} & \multicolumn{2}{c}{$\text{Image Recognition}^*$} & \multicolumn{3}{c}{Molecule Captioning} \\
    \cmidrule(rl){2-3}  \cmidrule(rl){4-6} 
    & \small{Molecules (Acc)} & \small{Reaction (Acc)} & \small{BLEU-2} & \small{BLEU-4} & \small{METEOR} \\
    \midrule
    \rowcolor{grey}\multicolumn{6}{c}{\textit{specialist models}} \\
    MolNextr (I)~\cite{chen2024molnextr} & 82.0 & -& -& -& -\\ 
    MolScribe (I)~\cite{qian2023molscribe} & 84.0 & -& -& -& -\\
    \midrule
    \rowcolor{grey}\multicolumn{6}{c}{\textit{generalist models}} \\
    ChemLLM-7B-Chat (S)~\cite{zhang2024chemllm} & - & - & 14.5 & 7.4 & 22.2 \\
    ChemDFM-v1.0 (S)~\cite{zhao2024chemdfm} & - & - & \underline{26.1} & \underline{20.6} & \underline{33.1} \\
    GPT-4O (S + I)~\cite{gpt-4v} & 0 & 0 & 7.0 & 0.4 & 15.0 \\
    Gemini-1.5-pro (S + I)~\cite{gemini} & 23.0 & 0 & 8.7 & 2.3 & 14.5 \\
    Qwen-VL-max (S + I)~\cite{qwenvl} & 0 & 0 & 5.9 & 0.1 & 15.0 \\
    LLaVA-v1.6-Vicuna-13B (S + I)~\cite{llava} & 0 & 0 & 4.4 & 0.0 & 10.7\\
    InternLM-XComposer2 (S + I)~\cite{internlmxcomposer} & 0 & 0 & 9.2 & 0.0 & 18.3 \\
    DocOwl-1.5-Omni (S + I)~\cite{hu2024mplug} & 0 & 0 & 10.0 & 1.6 & 17.9 \\
    \multirow{1}{*}{\textbf{ChemDFM-X (I)}} & 46.0 & 53.0 & 23.6 & 18.0 & 30.7 \\
    \multirow{1}{*}{\textbf{ChemDFM-X (S + I)}} & - & - & \textbf{27.2} & \textbf{21.7} & \textbf{34.3} \\
    \bottomrule
    \end{tabular}
    \caption{The Results of image recognition and molecule captioning tasks for the structural modalities. The content within the parentheses indicates the molecular representation modalities used for the corresponding model (S - SMILES, G - Molecular Graphs, I - Molecular Image). *: The SMILES notations are not included in the inputs of these tasks to prevent cheating. 
    Among the generalist models, the optimal results are \textbf{bolded}, and the second-best results are \underline{underlined}.}
    \label{tab:image-other}
\end{table*}

\begin{table*}
    \centering
    \begin{tabular}{lcccccc}
    \toprule
    Model & BACE & BBBP & Clintox & HIV & Tox21 & Avg \\
    \midrule
    ChemLLM-7B-Chat (S)~\cite{zhang2024chemllm} & \textbf{83.2} & 62.8 & 68.5 & 73.8 & 70.1 & 71.7 \\
    ChemDFM-v1.0 (S)~\cite{zhao2024chemdfm} & \underline{78.4} & \textbf{66.7} & \underline{89.9} & 73.6 & \textbf{79.8} & \underline{77.7} \\
    LLaVA-v1.6-Vicuna-13B (S + I)~\cite{llava} & 50.0 & 45.5 & 60.6 & 64.6 & 55.0 & 55.1\\
    InternLM-XComposer2 (S + I)~\cite{internlmxcomposer} & 46.8 & 51.7 & 50.3 & 49.6 & 41.7 & 48.0 \\
    DocOwl-1.5-Omni (S + I)~\cite{hu2024mplug}  & 44.4 & 53.4 & 50.0 & 57.0 & 48.4 & 50.6 \\
    \multirow{1}{*}{\textbf{ChemDFM-X (I)}} & 69.4 & \underline{65.7} & 73.3 & \underline{74.1} & 77.3 & 72.0 \\
    \multirow{1}{*}{\textbf{ChemDFM-X (S + I)}} & 76.9 & 65.1 & \textbf{93.5} & \textbf{75.3} & \underline{78.4} & \textbf{77.8} \\
    \bottomrule
    \end{tabular}
    \caption{The Results of molecule property prediction tasks for the image modality. We report the performance of the models in AUC-ROC, which stands for the Area Under the Curve of the Receiver Operating Characteristic. The content within the parentheses indicates the molecular representation modalities used for the corresponding model (S - SMILES, G - Molecular Graphs, I - Molecular Image).}
    \label{tab:image-molnet}
\end{table*}

\paragraph{Baselines.} Similar to structural modalities, we incorporate three types of baselines.
However, there is little previous work specifically targeted at chemical images. Therefore, we leverage multiple general-domain large vision-language models as the first type of baselines, including GPT-4V~\cite{gpt-4v}, Gemini~\cite{gemini}, Qwen-VL~\cite{qwenvl}, LLaVA-v1.6~\cite{llava}, InternLM-xcomposer2~\cite{internlmxcomposer}, and DocOwl-1.5-Omni~\cite{hu2024mplug}. The second type of baselines is the same chemical LLMs as those in structural modalities, namely ChemDFM and ChemLLM. 
The third type of baseline is the task-specific specialist model, GIT-Mol.

\paragraph{Results and Analysis.} The experimental results of the image modality are illustrated in Table~\ref{tab:image-other} and \ref{tab:image-molnet}. The results show that with the help of molecule images, ChemDFM-X can achieve comparable or better performance across all the tasks, showing that our ChemDFM-X possesses strong chemical image understanding prowess.

It is worth noticing that all the general-domain LMMs we have tested fail to complete the chemical tasks. To study the reason for this phenomenon, we manually examine all the outputs of the models. We find that since these general domain LMMs are not fine-tuned for specialized chemical tasks, the models lack sufficient chemistry knowledge to complete the tasks. For instance, in the molecule captioning task, some models misinterpret it as a SMILES generation task, yielding only the SMILES notation. Others, despite comprehending the task, lack the capacity to correlate molecules with their corresponding chemical knowledge, resulting in stubbornly generating molecule descriptions unrelated to the input molecules. In the molecule image recognition task, although the models can recognize a portion of the chemical information in the image, their deficiency in understanding SMILES rules hinders them from generating long SMILES notations, limiting task performance. In the reaction image recognition task, a portion of the models claim that they don't have the ability and refuse to answer, and other models generate invalid SMILES due to their lack of capacity for understanding the SMILES rules.

Another noteworthy result is the excellent performance of ChemDFM-X in the reaction image recognition task. Considering the length of a chemical reaction which is composed of numerous molecules, it is astonishing that the accuracy of reaction image recognition is even higher than that of single-molecule image recognition. We believe this precisely demonstrates the advantages and capabilities of cross-modality chemical LMMs, where ChemDFM-X uses the reaction knowledge it has acquired to help the reaction image recognition task. Specifically speaking, instead of single-molecule image recognition where the model needs to identify each and every bit of the molecule including the types of each atom and the connections between them, during reaction image recognition, the small mistakes when identifying each molecule may be tolerant because of the existence of the reaction context. These small mistakes may be corrected by the ChemDFM-X itself through analyzing the reaction. In other words, with the cross-modality comprehension capability, ChemDFM-X can utilize the reaction knowledge learned with SMILES representations to help implicitly correct minor mistakes during the reaction image recognition task, therefore resulting in much higher performances.

\subsection{Characterization Modalities}


\begin{figure}
    \centering
    \includegraphics[width=0.8\linewidth]{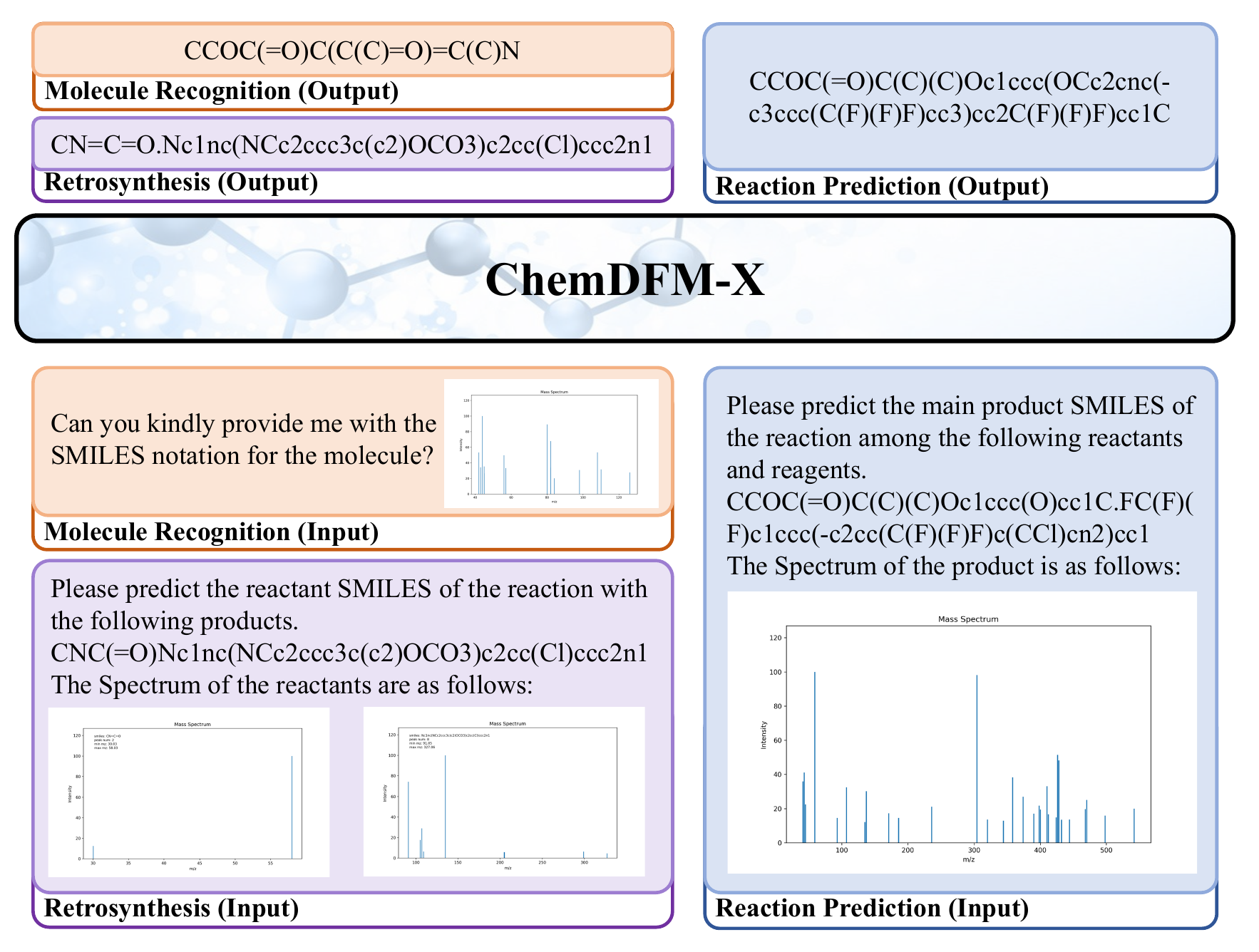}
    \caption{Evaluation tasks for characterization modalities}
    \label{figure:eval3}
\end{figure}

\begin{table*}[t]
    \centering
    \begin{tabular}{lccccc}
    \toprule
    \multirow{2.1}{*}{Model} & SR & \multicolumn{2}{c}{RP} & \multicolumn{2}{c}{Retrosynthesis} \\
    \cmidrule(rl){2-2} \cmidrule(rl){3-4} \cmidrule(rl){5-6}
    & \small{Acc} & \small{Acc} & \small{Validity} & \small{Acc} & \small{Validity} \\
    \midrule
    ChemLLM-7B-Chat (S)~\cite{zhang2024chemllm} & - & 18 & 94 & 10 & 91 \\
    ChemDFM-v1.0 (S)~\cite{zhao2024chemdfm} & - & \underline{49} & \underline{98} & 12 & \underline{99} \\
    \multirow{1}{*}{\textbf{ChemDFM-X (M)}} & 0 & - & - & - & - \\
    \multirow{1}{*}{\textbf{ChemDFM-X (R)}} & 1 & - & - & - & - \\
    \multirow{1}{*}{\textbf{ChemDFM-X (S + M)}} & - & 45 & 88 & \underline{29} & 84 \\
    \multirow{1}{*}{\textbf{ChemDFM-X (S + R)}} & - & \textbf{64} & \textbf{100} & \textbf{60} & \textbf{100} \\
    \bottomrule
    \end{tabular}
    \caption{The Results of multiple evaluation tasks, including Spectrum Recognition~(SR), Reaction Prediction~(RP), and Retrosynthesis, for the characterization modalities. The metric validity evaluates whether the output SMILES is valid. The content within the parentheses indicates the molecular representation modalities used for the corresponding model (S - SMILES, M - MS2 Spectra, R - IR Spectra). The optimal results are \textbf{bolded}, and the second-best results are \underline{underlined}.}
    \label{tab:spect-other}
\end{table*}

\paragraph{Evaluation Tasks.} Different from the aforementioned modalities which can only represent the known molecules, the characterization modalities are good at representing the unknown substances. The information they contained is mainly about partial properties and sub-structures and is usually implicit and scattered. Therefore, the data of characterization modalities are more suited for the unknown substance identification task.

Therefore, in our evaluation, we primarily focus on the spectrum identification tasks. Intuitively, the first group of tasks to evaluate models' spectrum identification capabilities is basic spectrum recognition, where the spectrum of a molecule is provided and the model is asked to identify the corresponding molecule based on that. However, during real chemical research, the unknown substances do not appear out of nowhere. Usually, the spectra are provided in the context of reactions, and the knowledge such as how the substance is synthesized and what reaction it can undergo is known. Therefore, it is vital and valuable for models to possess the capability to identify the molecule based on the molecular spectra along with the reaction related to the unknown molecule. To construct such tasks, we leverage the reaction prediction and retrosynthesis tasks used for structural modalities and ask models to generate the corresponding products or reactants based on the SMILES notation of the reaction and the spectra of the missing molecule.

An intuitive diagram demonstrating the input and output of the evaluation tasks for characterization modalities is shown in Figure~\ref{figure:eval3}.

\paragraph{Baselines.} The development of spectrum identification is relatively early-stage. Currently, there have still not been public-available standard baselines and benchmarks. Therefore, we only compare our ChemDFM-X model with the conventional chemical LLMs, ChemDFM, and ChemLLM, which only comprehend text and SMILES.

\paragraph{Results and Analysis.}The experimental results of the characterization modalities are illustrated in Table~\ref{tab:spect-other}. 
It is noteworthy that although the top 1 accuracy of the spectrum recognition task has near-zero performance, the performance leaps significantly when additional reaction information is involved. Note that the resulting performances (64\% in the reaction prediction task and 60\% in the retrosynthesis task) are far higher than those of the origin tasks where only reaction SMILES are provided with no spectra. Therefore, what is happening here is not a strong modality helping a weak modality, but a collaboration between modalities. From a chemical perspective, the spectrum modalities can provide SMILES modality hints related to the structure and composition information of the missing substances in the reactions, while SMILES modality can trim incorrect options. Through this cross-modality collaboration, ChemDFM-X is able to achieve much better performances greater than the sum of its parts. In short, ChemDFM-X successfully gains the ability to understand and analyze different spectra and is able to combine the information from different modalities to achieve higher performance.

%% file: 5.conclusion.tex
\section{Conclusion}

This paper proposed ChemDFM-X, a large multimodal model for chemistry. ChemDFM-X is a generalist model that has the ability to understand five of the most commonly used modalities in the field of chemistry, including structural modalities, image modalities, and characterization modalities.
The evaluation result shows that ChemDFM-X possesses the capabilities to comprehend the chemical data in all five non-text modalities. With the help of multi-modality inputs, ChemDFM-X is capable of exploiting cross-modality information and knowledge, especially the spectrum input introducing experimental observations lacking in other models. This assists ChemDFM-X in outperforming other generalist models in a series of common chemical tasks and demonstrates the practical value of ChemDFM-X in real chemistry research. It also has the potential for dealing with inputs of multiple different modalities simultaneously, which is powerful in reaction-related tasks and will be further tested in our future work.